\documentclass{article}

\PassOptionsToPackage{numbers, compress}{natbib}
\usepackage[preprint]{neurips_2024}
\usepackage{amsfonts}   
\usepackage{amsmath}
\usepackage{amssymb}
\usepackage{graphicx}
\usepackage[utf8]{inputenc} 
\usepackage[T1]{fontenc}    
\usepackage{hyperref}       
\usepackage{url}            
\usepackage{booktabs}       
\usepackage{amsfonts}       
\usepackage{nicefrac}       
\usepackage{microtype}      
\usepackage{xcolor}         

\begin{document}

\title{Revisiting Neighborhood Aggregation in Graph Neural Networks
for Node Classification using \\ Statistical Signal Processing
}

\author{Mounir Ghogho, {\normalfont {\em IEEE Fellow}} \vspace{0.5cm} \\ 
International University of Rabat, TICLab, Morocco \\
\texttt{mounir.ghogho@uir.ac.ma} 
}

\maketitle

\begin{abstract}
We delve into the issue of node classification within graphs, specifically reevaluating the concept of neighborhood aggregation, which is a fundamental component in graph neural networks (GNNs). Our analysis reveals conceptual flaws within certain benchmark GNN models when operating under the assumption of {\em edge-independent node labels}, a condition commonly observed in benchmark graphs employed for node classification. Approaching neighborhood aggregation from a statistical signal processing perspective, our investigation provides novel insights which may be used to design more efficient GNN models. 
\end{abstract}

\section{Introduction}
\label{Intro}

Consider an attributed graph $\mathcal{G}=(\mathcal{V},\mathcal{E}, \mathcal{X})$ where $\mathcal{V}$ is the set of nodes,  $\mathcal{E}$ is the set of edges, i.e. pairs of connected nodes, and $\mathcal{X}=\{ \boldsymbol{x}_i, i\in \cal{V} \}$ is the set of node feature vectors,  of dimension $F$ each. Let $\mathcal{N}_i$ denote the set of immediate neighbors of node $i$, and let $d_i$ denote its degree, defined as $d_i=|\mathcal{N}_i|$, where $|.|$ denotes the cardinality operator. We assume that each node belongs to one of $M$ classes and we denote the prior probability of class $m$ by $\pi_m$. Some of the nodes are labeled (i.e. the class they belong to is known), and the goal of node classification is to predict the labels of unlabeled nodes. 
We consider the case where the nodes are classified independently, which is the approach taken in graph neural network (GNN) based methods.

We address node classification under the {\em edge-independent node labels} (EINL) assumption, which entails that the graph structure does not play any role in the ground node label determination. This assumption holds true under the condition that the labels assigned to the nodes are intrinsically linked to the respective node objects and remain constant irrespective of alterations in the interconnections between nodes. This circumstance is prevalent in a considerable array of graphs commonly employed as benchmarks for node classification, such as citation graphs, internet graphs, protein-protein interaction graphs, and social networks where the node labels do not change due to social influence, although changes to the labels due to the latter may still accommodate the EINL assumption if the node features describing the online activities of the users adapt to the evolving labels.

We formulate the EINL assumption mathematically by assuming that the true label of a node depends solely on the statistical distribution of its feature vector. We thus make the following mathematical assumption:

({\bf A1}) {\em The feature vectors of nodes belonging to the same class are randomly drawn from the same multivariate distribution.} 

Let $D_m(\boldsymbol{x}_i)$ denote the distribution of node feature vector $\boldsymbol{x}_i$ when the node belongs to class $m$. The EINL assumption implies that if the node feature distributions corresponding to different classes do not overlap, label prediction would be error-free. 
It is noteworthy to emphasize that the degree of expressiveness inherent in a node's representation, specifically within its feature vector, significantly influences the ease of the classification task. 

The extent of overlap among the class distributions directly correlates with the potential for misclassification errors \cite{Seoane2023}.
Leveraging the graph structure, particularly through label-to-label correlations, presents an opportunity to mitigate such misclassification errors.

\begin{figure}[h!]
\hspace{-16mm}
\includegraphics[height=1.4in,width=6.7in]
{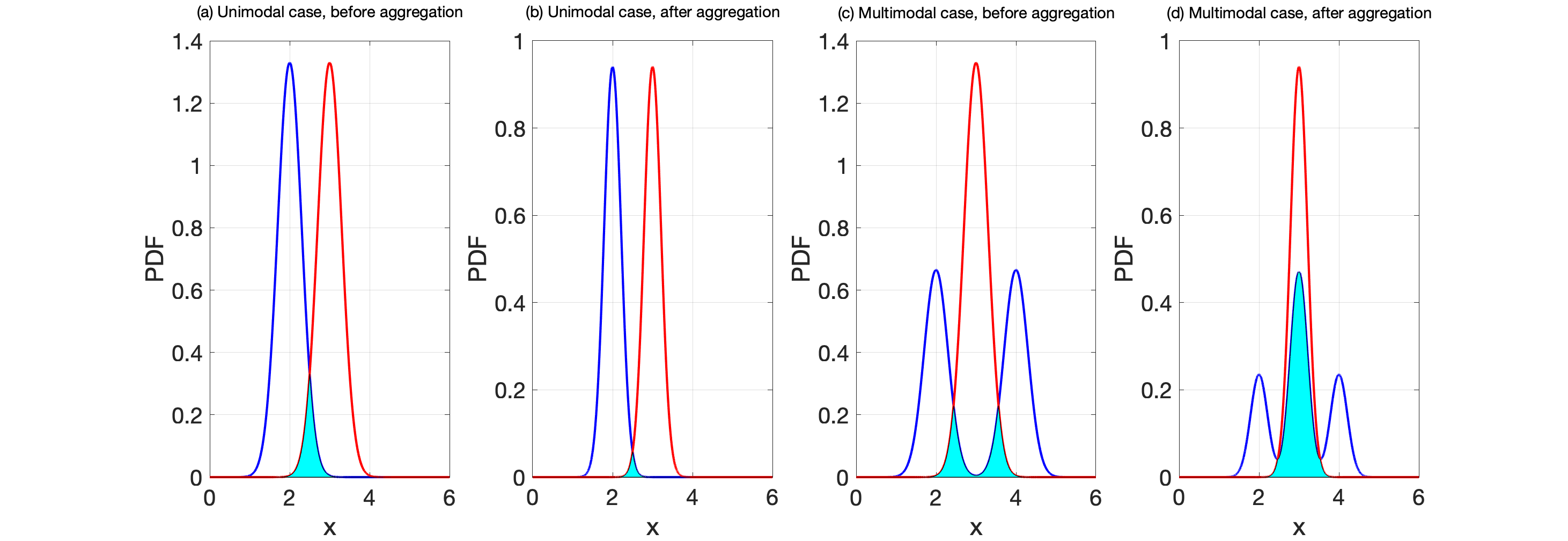}
 \vspace{-4mm}
\caption{\small{Candidate distributions before and after  neighborhood aggregation, $\boldsymbol{x}_i+\boldsymbol{x}_j$, assuming pure homophily and $F=1$, in the unimodal and multimodal feature  distribution cases.}}
    \label{fig:pdf_d1}
\end{figure}

To harness label-to-label correlations, GNN typically employs iterative message passing where the messages are  nodes' representation vectors, which are initialized by the node feature vectors \cite{Bronstein}\cite{Veličković2022}.  
More fundamentally, GNNs leverage label-to-label correlations through neighborhood aggregation, i.e. the feature vector of a node is aggregated with those of its neighbors \cite{Hamilton-book}. This concept is rooted in the underlying assumption of homophily, which suggests that connected nodes tend to belong to the same class \cite{McPherson}.  
Using our statistical perspective, the effectiveness of neighborhood aggregation can be attributed to the fact that the candidate distributions of the nodes' representation vectors resulting from this aggregation exhibit less overlap than those of the original feature vectors, provided that these are {\em unimodal}. Indeed, with unimodal distributions, which peak at a certain value of the feature vector and taper off on both sides, and assuming pure homophily, the distribution of $\boldsymbol{x}_i+\boldsymbol{x}_j$ will be the convolution of the distribution of $\boldsymbol{x}_i$ with itself.  This convolution accentuates the central peak and reduces the tails. This leads to a narrower shape in the resulting distribution, which hence reduces the overlap between the candidate distributions, as depicted in subplots (a) and (b) of  Figure \ref{fig:pdf_d1}.

The following remarks encapsulate the core motivation underpinning this work.\\
{\bf Remark 1}. Under the EINL assumption, when performing weighted sum aggregation \footnote{This aggregation, which is  investigated in Section \ref{WSAsection}, is a special case of linear aggregation.} in a homophilic graph, each neighbor must {\em at most} have the same weight as the node itself, as otherwise, undue preference would be conferred upon the neighbor. Furthermore, if the distributions of the feature vectors are unimodal, greater homophily, signifying higher confidence in a neighbor sharing the same class as the focal node, should result in the neighbor's weight approaching that of the focal node.  $\hfill \square$ \\
{\bf Remark 2}. The widely adopted  GraphConvolutional Network (GCN) \cite{GCN} performs reasonably well on benchmark homophilic graphs by harnessing neighborhood aggregation. However, in light of Remark 1, the approach to aggregation employed seems counterintuitive for node classification under the EINL assumption. Indeed,  the weight assigned to the focal node, say node $i$, is $1/d_i$ and those assigned to the neighbors, $j\in\mathcal{N}_i$, are $1/\sqrt{d_id_j}$.
Hence, the weight assigned to neighbor $j$ would be larger than that assigned to the focal node $i$ when $d_j<d_i$. This can be attributed to the fact that GCN was originally conceived for a different purpose, specifically, to approximate spectral graph convolution.  $\hfill \square$ \\ 
{\bf Remark 3}. In GNN research, much attention has been paid to devising algorithms that are as expressive as the Weissfeiler Lehman (WL) isomorphism test \cite{Xu2019}. Further, motivated by the observation that WL algorithm fail to tell apart some basic regular graphs, some researchers have proposed GNN models that are more expressive than WL algorithm. One research direction has been to design GNN models that are as expressive as higher-dimensional WL algorithms, where sets of nodes instead of single nodes are used in the WL aggregation process \cite{Maron2019}. 
However, under the EINL assumption, the node labels are not dependent on the graph structure and thus, ensuring expressivity of the WL type is not required for node classification. One can easily find a GNN which is not as powerful as the WL algorithm and yet more accurate in terms of classification performance. Therefore, utilizing WL isomorphism test as inspiration in the development of GNN, such as the Graph Isomorphism Network (GIN) \cite{Xu2019}, is not theoretically justified for node classification when the graph conforms to the EINL condition.$\hfill \square$\\
{\bf Remark 4}. If at least one of the feature distributions is multi-modal, {\em linear} aggregation, as in GCN and GIN, may compromise performance, even in the presence of strong homophily within the graph. To illustrate this, consider the case of pure homophily and two equally probable classes, and assume that the feature distribution in Class 1 is a Gaussian mixture, i.e. $D_1(\boldsymbol{x}_i)=0.5{G}(\boldsymbol{x}_i | \boldsymbol{\mu}_1,{\bf C})+0.5 {G}(\boldsymbol{x}_i | \boldsymbol{\mu}_2,{\bf C})$,\footnote{$G(\boldsymbol{x}_i | \boldsymbol{\mu},{\bf C})$ denote a multivariate Gaussian distribution with mean $\boldsymbol{\mu}$ and covariance matrix {\bf C}.} and that $D_2(\boldsymbol{x}_i)={G}(\boldsymbol{x}_i |(\boldsymbol{\mu}_1+\boldsymbol{\mu}_2)/2,{\bf C})$. Linearly aggregating the feature vector of a node and that of a neighbor $\boldsymbol{x}_j$ using $\boldsymbol{z}_i = (\boldsymbol{x}_i +\boldsymbol{x}_j$)/2, would result in the following candidate distributions: $0.25{G}(\boldsymbol{z}_i |\boldsymbol{\mu}_1,{\bf C}/2)+0.5{G}(\boldsymbol{z}_i | (\boldsymbol{\mu}_1+\boldsymbol{\mu}_2)/2,{\bf C}/2)+0.25{G}(\boldsymbol{z}_i | \boldsymbol{\mu}_2,{\bf C}/2)$ for Class 1 and ${G}(\boldsymbol{z}_i | (\boldsymbol{\mu}_1+\boldsymbol{\mu}_2)/2,{\bf C}/2)$ for Class 2. Hence, in the above scenario and unlike the case of unimodal feature distributions, linear aggregation increases the overlap between the candidate distributions with respect to the graph agnostic classifier instead of reducing it even if the graph is purely homophilic; see subplots (c) and (d) of Figure \ref{fig:pdf_d1}. The notable superiority of GNNs based on linear neighborhood aggregation, such as GCN and GIN, over graph-agnostic classifiers when applied to benchmark homophilic graphs for node classification suggests that in benchmark graphs the feature distributions tend to be unimodal.
$\hfill \square$ \\

In this paper, we revisit neighborhood aggregation and investigate the performance of node classification from a statistical perspective. 
For the sake of analytical tractability, we restrict our consideration to GNNs with a single layer, indicating the aggregation of information solely from immediate neighbors. Although this setting is restrictive, it facilitates valuable insights into the ways in which graph structure can enhance node classification. Further, it has been demonstrated that the nonlinearity between GNN layers is not deemed crucial, with the primary advantages attributed to neighborhood aggregation \cite{Wu2019}. This observation suggests the possibility of eliminating the nonlinear transition functions between each layer, retaining solely the final softmax for the generation of probabilistic outputs. Our analysis can therefore be extended to this multilayer GNN setting in a straightforward manner by directly aggregating neighbors of different orders. This is briefly discussed in Section \ref{extensions}.

In certain graphs, the magnitude of the feature vector within each class may exhibit significant variations across nodes. In such instances, the assumption of fixed distributions may be approximately validated by normalizing the feature vectors of all nodes in the graph to possess the same norm.

Finally, it is worth emphasizing that the investigation conducted in this paper is delimited to graphs wherein the EINL condition is satisfied.  In instances where the graph features edge-{\em dependent} node labels, meaning that node labels are contingent upon graph edges, the formulation of the node classification problem necessitates a distinct approach. The examination of this particular issue in the context of hypergraphs has been explored in \cite{Choe2023}.
\section{Preliminaries}
\subsection{Graph modeling}
\begin{figure}
 \hspace{-15mm}
\includegraphics[height=3.4in,width=6.7in]
{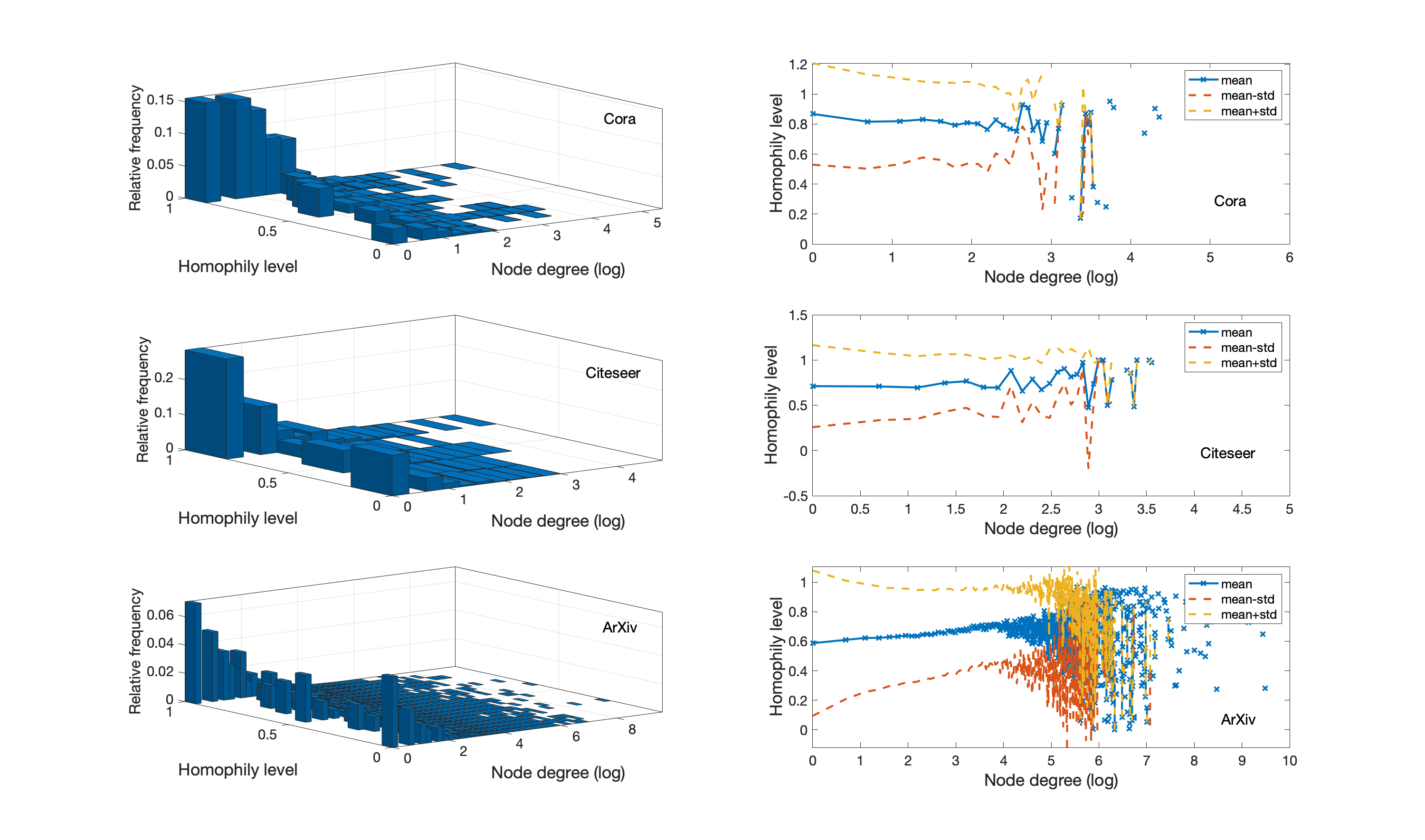}
 \vspace{-4mm}
\caption{\small{Homophily level versus node degree.}}
    \label{fig:homophilydegre}
\end{figure}

Real-world graphs are often homophilic. Strong homophily translates to high label-label correlations, thereby substantially enhancing performance compared to graph-agnostic classification approaches. Let $\kappa_i=(1/d_i)\sum_{j\in\mathcal{N}_i}\mathcal{I}(y_j=y_i)$ denote the per-node homophily level at node $i$, where $\mathcal{I}(y_j=y_i)$ is the indicator function, i.e. $\mathcal{I}(A)=1$ if $A$ is true and zero otherwise; and let $p_{h}(d)$ denote the average per-node-degree homophily level, i.e. $p_h(d)=\frac{1}{|\mathcal{V}_d|} \sum_{i\in  \mathcal{V}_d} \kappa_i$, where $\mathcal{V}_d$ is the subset of nodes having degree equal to $d$. 
Figure \ref{fig:homophilydegre} depicts the histrograms and first and second-order statistics of the node-level homophily for different node outdegrees for citation graphs Cora, Citeseer and ArXiv. The average homophily levels of the three graphs are respectively 0.8, 0.76 and 0.63 respectively.
The figure shows that for Cora and CiteSeer, the average per-node-degree homophily level does not exhibit a discernible consistent trend with the node degree. For the ArXiv graph, a subtle gradual increase in the average homophily is discernible for node degrees ranging from 1 to approximately 50 (or 0 to 4 in log scale). The pronounced fluctuations in the average per-node-degree homophily for larger node degrees primarily stem from the limited count of nodes possessing those degrees. 
Based on the above observations and to simplify the analysis, we model the label-to-label correlations in the graph using the following reasonable and commonly used Markov-like assumption \cite{Qu2019,Hafidi2023}:

\noindent{\bf (A2)} {\em for any non-zero-degree node, say node $i$, of class $m$, the probability that a randomly selected neighbor, say node $j$, belongs to class $\ell$ is constant throughout the graph; we denote this probability by $p_{m,\ell}$, i.e.
\begin{equation}
    p_{m,\ell}\triangleq {\rm Pr}\left( y_j=\ell | y_i=m\right) \nonumber 
\end{equation} 
}

When $\ell=m$, the above probabilities are referred to as the per-class homophily levels. A graph is said to be {\em homophilic} if $p_{m,m}>0.5$, $\forall m$. The average homophily level can be expressed in terms of the per-class homophily levels as $p_h=\sum_{m}\pi_m p_{m,m}$. A consequence of assumption (A2) is that the label-to-label correlations are independent of the nodes' degrees. Graphs under assumption (A2) can be generated using the degree-corrected version of the stochastic blockmodel \cite{Karrer}. This model generalizes the conventional stochastic blockmodel by allowing to specify the expected degrees of the different nodes.  

We assume that the distributions of the feature vectors as well as the class transition probabilities, the $p_{m,\ell}$'s, and the prior probabilities, the $\pi_m$'s, are known. In  practice, these have to be estimated using the feature vectors of the labeled and unlabeled nodes as well as the graph edges.
Here, we disregard the resulting estimation errors to underscore the potential of graph connections in enhancing node classification performance.

\subsection{Node feature distribution modeling}
In light of Remark 4, for linear neighborhood aggregation to be always beneficial, we make the following assumption.

({\bf A3}) {\em The node feature distributions, the $D_m(\boldsymbol{x})$'s, are unimodal. 
}

Let the mean vectors and covariance matrices associated with the above-mentioned distributions be denoted as $\boldsymbol{\mu}_m$ and ${\bf C}_m$, respectively, with $m=1,\cdots,M$. To better elucidate the effects of neighborhood aggregation and streamline our analytical derivations, we will consider the following special scenario of assumption (A3).

{\em {\bf Homoscedastic Gaussian (HG) scenario}: 
The $D_m(\boldsymbol{x})$'s are Gaussian distributions with different mean vectors but the same covariance matrix ${\bf C}$ which is assumed to be full-rank. 
}

The assumption of homoscedasticity, where the class covariance matrices are identical, simplifies the derivations. In scenarios where the class covariance matrices differ, the assumption may still be applied by utilizing the pooled covariance matrix, which represents a weighted average of the individual class covariance matrices. Additionally, the Gaussian assumption further facilitates analytical derivations. While this assumption may not be appropriate for some real-world graphs, particularly when node features are binary random variables, linear processing of feature vectors, which is inherent in multilayer perceptron (MLP) models, can be approximated by Gaussian variables due to the central limit theorem, provided that the feature dimension $F$ is sufficiently large. Therefore, the Gaussian assumption on the feature vectors is not overly restrictive.

\subsection{Graph-agnostic classifier}\label{GACsubsection}
 
The single-layer MLP-based classifier is
$\hat{y}_i=\arg\max_m (\boldsymbol{w}_{m}^T\boldsymbol{x}_i+b_{m})
$, where the $\boldsymbol{w}_{m}$'s and the $b_m$'s are respectively ($F\times 1$) vectors and scalars to be optimized. Under our HG scenario, this classifier with $\boldsymbol{w}_{m}={\bf C}^{-1}\boldsymbol{\mu}_m$ and $b_m=-\frac{1}{2}\boldsymbol{\mu}_m^T{\bf C}^{-1}\boldsymbol{\mu}_m+\log(\pi_m)$ coincides with the optimum Bayesian classifier,
and the corresponding misclassification error probability is lower-bounded as follows:
\begin{equation}    
e\triangleq {\rm Pr}(\hat{y}_i\ne y_i)\le\sum_{m=1}^M \pi_m \sum_{\ell\neq m} Q\left(\frac{\sqrt{\gamma_{0}(m,\ell)}}{2}+\frac{1}{\sqrt{\gamma_0(m,\ell)}}\log\frac{\pi_m}{\pi_\ell}\right)
    \label{e0m}
\end{equation}
where $\gamma_0(m,\ell)=(\boldsymbol{\mu}_m-\boldsymbol{\mu}_\ell)^T{\bf C}^{-1}(\boldsymbol{\mu}_m-\boldsymbol{\mu}_\ell)$ and $Q$ is the $Q$-function \footnote{$Q(a)=\int_a^{+\infty}(1/\sqrt{2\pi})\exp(-t^2/2)dt$.}; see proof in Appendix A.  The above inequality becomes an equality in the binary case, i.e. $M=2$. 
It is worth noting that the $\gamma_0(m,\ell)$'s play a key role in classification performance. They quantify the degree of separability between the different classes in the Gaussian case. In the case of binary classification, $\gamma_0(1,2)$ (or simply $\gamma_0$ as there is only one such coefficient in the binary case)  is known as the {\em deflection} coefficient in the signal processing community \cite{Picinbono}, where  
$\gamma_0=\|\boldsymbol{\mu}_1-\boldsymbol{\mu}_2\|^2/\sigma^2$ because ${\bf C}$ was set to $\sigma^2{\bf I}_F$, with ${\bf I}_F$ denoting the $(F\times F)$ identity matrix.
Since the error probability is a monotonically decreasing function of $\gamma_0$, the latter is often used as a measure of performance. In this paper, we refer to the $\gamma_0(m,\ell)$'s as the graph-agnostic deflection coefficients.

\section{Weighted sum aggregation}\label{WSAsection}
We first consider the following weighted sum aggregation (WSA):
\begin{equation}
\boldsymbol{z}_i=\alpha_{i,i}\boldsymbol{x}_i+\sum_{j\in{\mathcal N}_i} \alpha_{i,j}\boldsymbol{x}_j \label{ave0}
\end{equation}
GNN models based on the above aggregation are  referred to in the GNN literature as convolutional GNN \cite{Bronstein}. Examples of such GNNs include GCN \cite{GCN}, where $\alpha_{i,i}=1/d_i$ and $\alpha_{i,j}=1/(\sqrt{d_i d_j}), \forall j\in \mathcal{N}_i$, and GIN \cite{Xu2019}, where $\alpha_{i,i}=1+\epsilon$ and $\alpha_{i,j}=1, \forall j\in \mathcal{N}_i$. Next, we investigate the optimization of weighting coefficients to maximize statistical classification performance.

{\em The case of purely homophilic graphs}: in this case, under assumption (A3), optimal classification performance is obtained when the neighbors are assigned the same weight as the focal node.
Indeed, considering our HG scenario, the mean of $\boldsymbol{z}_i$ is $(\alpha_{i,i}+\sum_j\alpha_{i,j}) \boldsymbol{\mu}_m$ when $y_i=m$; the covariance matrix is $(\alpha_{i,i}^2+\sum_j\alpha_{i,j}^2){\bf C}$, since the feature vectors of different nodes are assumed to be independent. The deflection coefficients for node $i$ become $\gamma_{d_i}(m,\ell)=[(\alpha_{i,i}+\sum_j\alpha_{i,j})^2/(\alpha_{i,i}^2+\sum_j\alpha_{i,j}^2)] \gamma_0(m,\ell)$. It is straightforward to show that these coefficients are maximized with $\alpha_{i,j}=\alpha_{i,i}, \forall j\in \mathcal{N}_i$. 
The resulting deflections are  
$\gamma_{d_i}(m,\ell)=(d_i+1)\gamma_0(m,\ell)$; factor $d_i+1$ can be interpreted as the {\em graph gain} for classifying node $i$ when using first-order neighbors \footnote{The reference of the graph gain is 1, which corresponds to the graph-agnostic classifier}. 
This gain can be attributed to a reduction in the covariance matrix of the nodes' representation by a factor of $(d_i+1)$, while the means remain unchanged.
The optimised single-layer MLP-based classifier is under the HG scenario the same as in the graph-agnostic classification case after substituting $\boldsymbol{z}_i$ for $\boldsymbol{x}_i$ and using the following bias term: $b_m=-(1/2)\tilde{\alpha}_{i,i}(1+d_i)\boldsymbol{\mu}_m^T{\bf C}^{-1}\boldsymbol{\mu}_m+\tilde{\alpha}_{i,i}\log\pi_{m,d_i}$ with $\pi_{m,d_i}$ denoting  the node degree-dependent a priori probabilities, in cases where the degree distribution is class-dependent. It is worth pointing out that the bias now is node-degree dependent, even if we ignore the term related to the  $\pi_{m,d_i}$'s, unless we normalise $\boldsymbol{z}_i$ by setting $\alpha_{i,i}=1/(d_i+1)$.  This sheds light on the debate regarding whether or not $\boldsymbol{z}_i$ should be normalized \cite{Xu2019}. Indeed, when $\boldsymbol{z}_i$ is not normalized, as in GIN, which was advocated to preserve the injectivity of the GNN model and thus make it as powerful as the WL algorithm, using fixed biases in the MLP classifier may undermine performance. 

\begin{figure}[!htb]
\minipage{0.5\textwidth}
  \includegraphics[width=\linewidth]{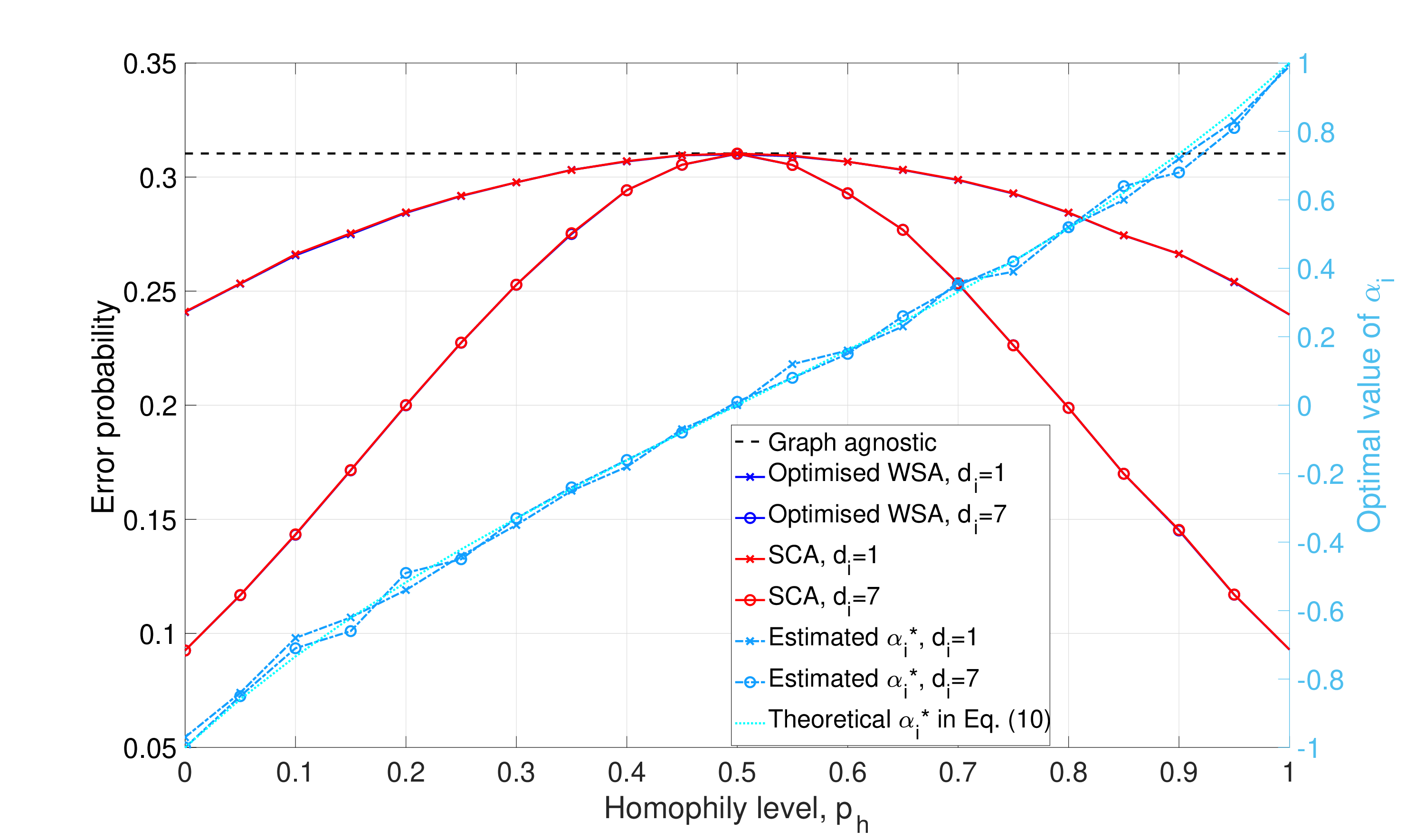}
\endminipage\hfill
\minipage{0.5\textwidth}
  \includegraphics[width=\linewidth]{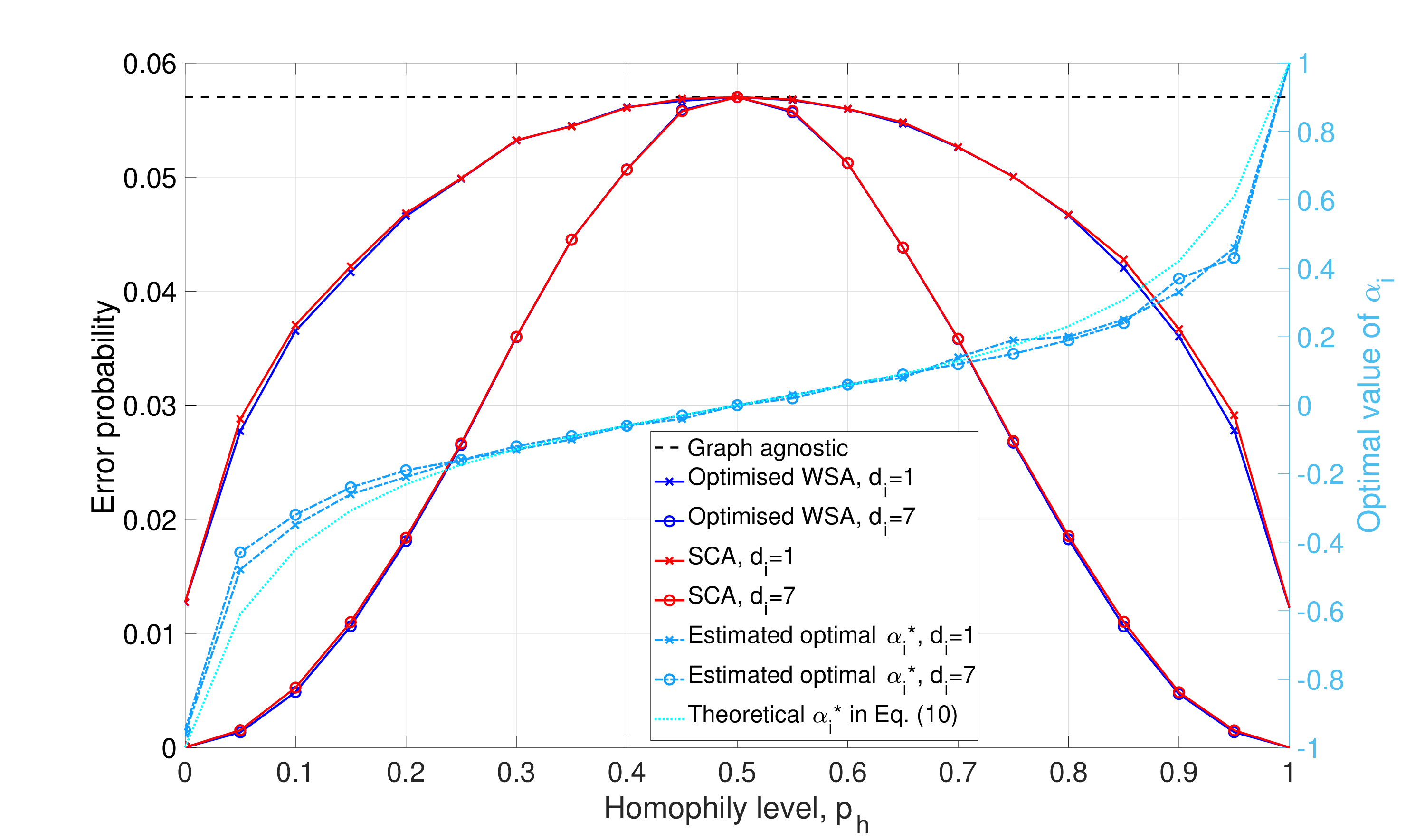}
\endminipage\hfill
\caption{Classification error probability of WSA and SCA and the optimum value of $\tilde{\alpha}_i$ for WSA versus $p_h$, assuming Special case 1; $\gamma_0=1$ (left) and $\gamma_0=10$ (right).}
\label{perf_M2}
\end{figure}

\begin{figure}[!htb]
    \centering

\minipage{0.5\textwidth}
\includegraphics[width=\linewidth]{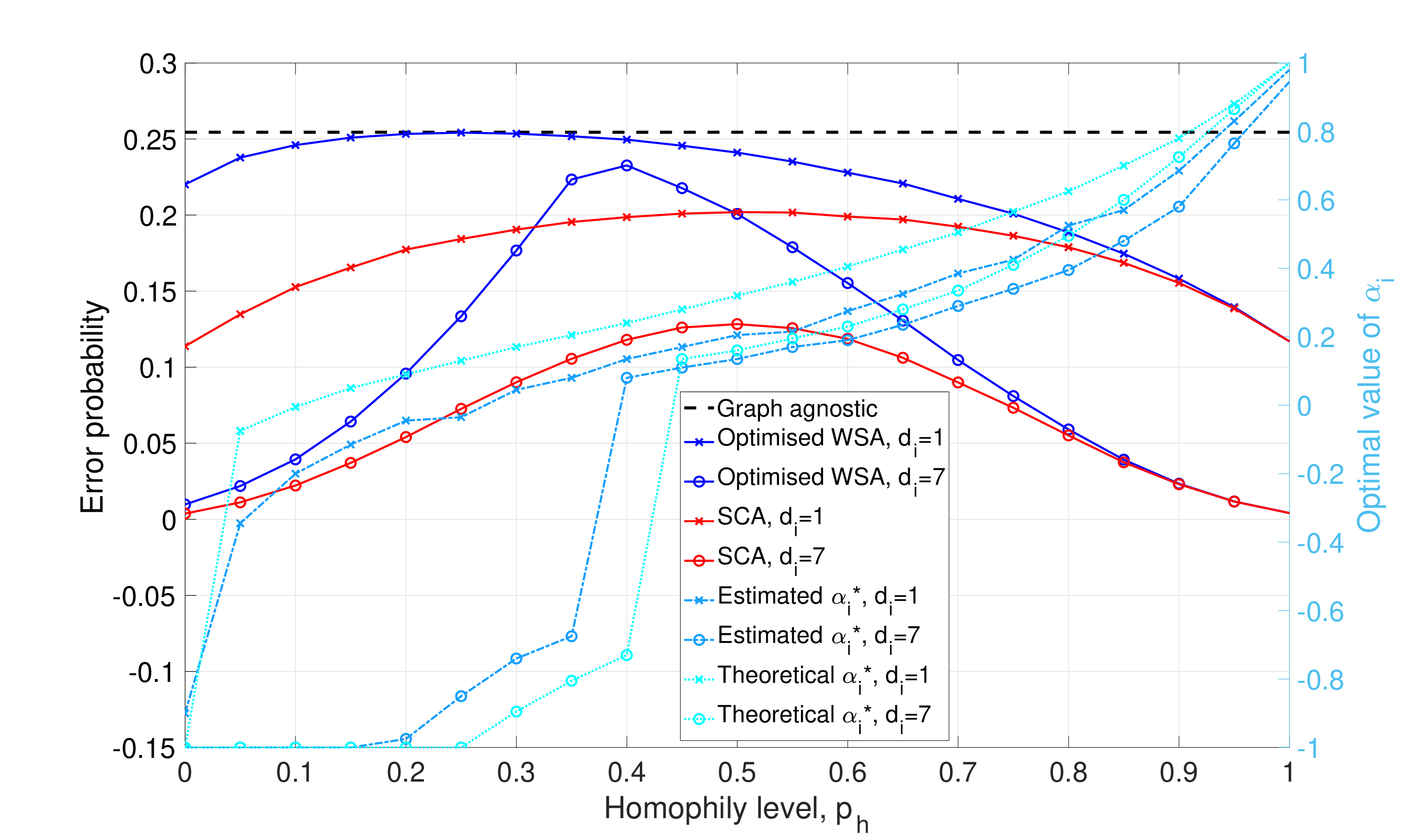}
\endminipage\hfill  
\minipage{0.5\textwidth}  
\includegraphics[width=\linewidth]{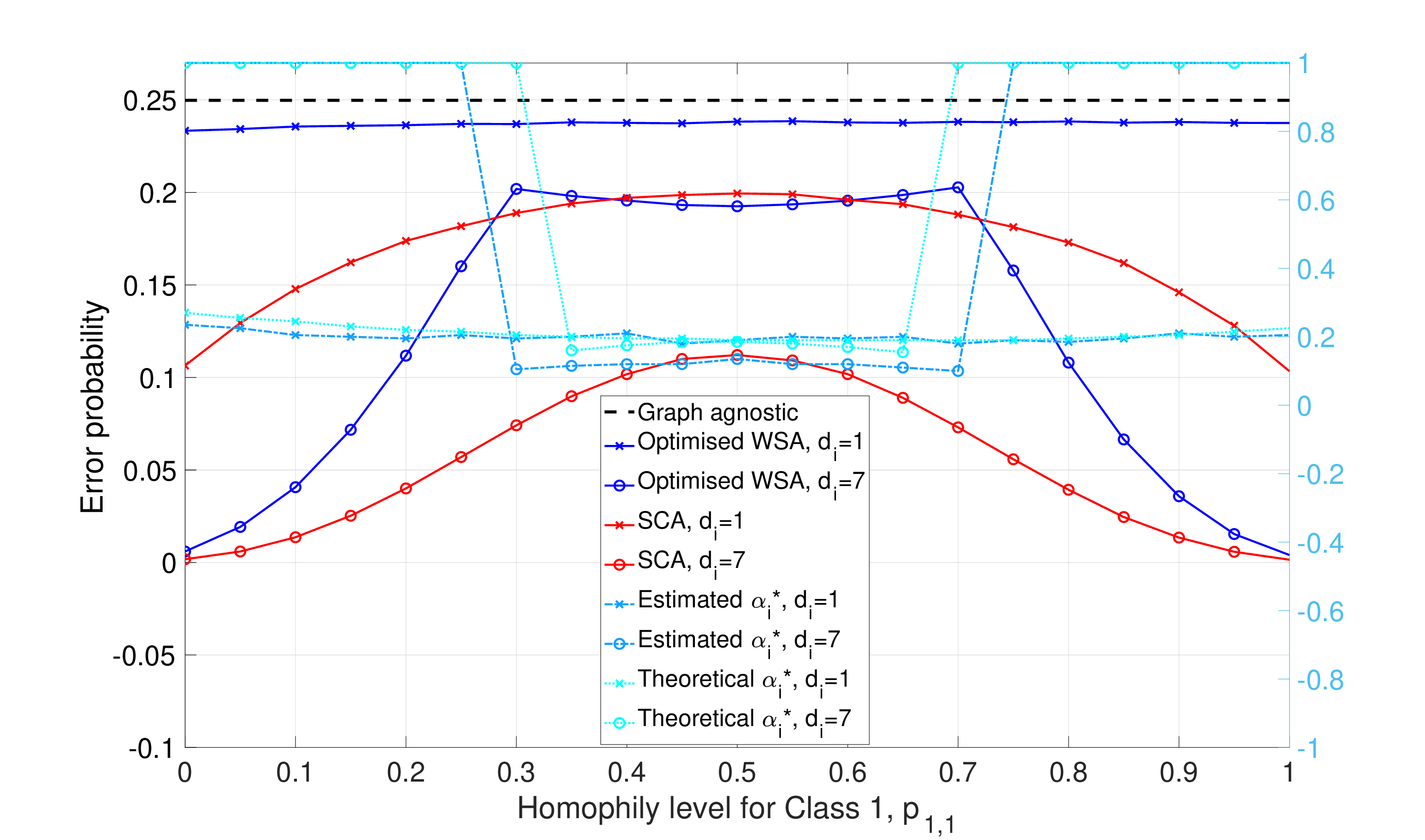}
\endminipage\hfill  
\caption{Classification error probability of WSA and SCA and the optimum value of $\Tilde{\alpha}$ versus the homophily level, $p_h$, assuming $M=4$, $\gamma_{0}(m,\ell)=4,\forall m\neq \ell$, in Special case 2 (left) and Special case 3 (right).}
    \label{perf_M4_gmin4}
\end{figure}

{\em The case of purely heterophilic graphs}: consider the case of two classes, i.e. $M=2$, which is common for heterophylic graphs (e.g. online transaction networks where nodes are either fraudsters or customers with the former being more likely to build connections with the latter instead of other fraudsters \cite{Pandit2007}, and dating networks where connections are more likely to occur people of different genders \cite{Zhu2021}). Pure heterophily implies in this case that $p_{1,1}=p_{2,2}=0$ and $p_{1,2}=p_{2,1}=1$. 
Considering our HG scenario, the mean of $\boldsymbol{z}_i$ is $\alpha_{i,i}\boldsymbol{\mu}_1+\sum_j\alpha_{i,j}\boldsymbol{\mu}_2$ if $y_i=1$ and $\alpha_{i,i}\boldsymbol{\mu}_2+\sum_j\alpha_{i,j}\boldsymbol{\mu}_1$ if $y_i=2$; the covariance matrix is ${(\alpha_{i,i}^2+\sum_j \alpha_{i,j}^2)}{\bf C}$ in both cases. It is straightforward to show that the corresponding deflection coefficient 
attains its maximum value with  $\alpha_{i,j}=-\alpha_{i,i}, \forall j\in \mathcal{N}_i$. The resulting deflection coefficient and graph gain coincide with those derived for the purely homophilic scenario with $\alpha_{i,j}=\alpha_{i,i}$; this is illustrated in Figure \ref{perf_M2}. It is worth noting that GCN and GIN may perform poorly in this setting due to the fact that the corresponding weighting coefficients are all positive (see for example \cite{Zhu2021} and \cite{Zhu2023} for the limitations of GCN in heterophilic graphs). In cases where $M > 2$, the graph-induced gain when using the aggregation in Eq. (\ref{ave0}) may not be as high as in the purely homophilic graph case, particularly for small values of $d_i$, as illustrated in Figure \ref{perf_M4_gmin4}.  

Practical graphs are neither purely homophilic nor purely heterophilic. Thus, under the EINL assumption, for homophilic graphs, the weight assigned to each neighbor of node $i$ must reflect the 'likelihood' that this neighbor belongs to the same class as node $i$. More generally, the weight assigned to a neighbor in absolute value must reflect the certainty about the class it belongs to given that of the focal node. In what follows, in order for the mean of $\boldsymbol{z}_i$ to be equal to that of $\boldsymbol{x}_i$ when neighbors are not used, without loss of generality, we set $\alpha_{i,i}=1$.

Under assumption (A2), the probability that a given neighbor belongs to the same class as node $i$ is independent of the neighbor's degree. We can therefore infer that under this assumption, the coefficients assigned to the different neighbors in Eq. (\ref{ave0}) should be equal to each other, i.e. $\alpha_{i,j}=\tilde{\alpha}_{i}, \forall j\in \mathcal{N}_i$. 
The aggregation in (\ref{ave0}) becomes  
\begin{equation}
\boldsymbol{z}_i=\boldsymbol{x}_i+\tilde{\alpha}_{i}\sum_{j\in{\mathcal N}_i} \boldsymbol{x}_j  \label{ave1}
\end{equation}

Let $n_{i,q}$ represent the count of neighbors of node $i$ belonging to Class $q$. Conditioned on these counts, $\boldsymbol{z}_i$ in Eq. (\ref{ave1}) is, when $y_i=m$ and under the HG scenario, Gaussian with mean given by: \begin{equation}    \boldsymbol{\nu}_m(\boldsymbol{n}_i)= (1+\tilde{\alpha}_in_{i,m}) \boldsymbol{\mu}_m+\tilde{\alpha}_i\sum_{q\neq m} n_{i,q} \boldsymbol{\mu}_q,
    \label{num}
\end{equation}
where $\boldsymbol{n}_i=\{n_{i,1},\cdots,n_{i,M}\}$ with $\sum_{q=1}^{M} n_{i,q}=d_i$, and covariance matrix given by  $(1+ \tilde{\alpha_i}^2d_i){\bf C}$.
When not conditioned on $\boldsymbol{n}_i$, the distribution of $\boldsymbol{z}_i$ given $y_i$ adheres to a Gaussian mixture model. Under assumption (A2), its unconditional mean and covariance matrix when $y_i=m$ are found to be
\begin{eqnarray}    
\Bar{\boldsymbol{\nu}}_{m,d_i}&=& \boldsymbol{\mu}_m+ \tilde{\alpha}_id_i \bar{\boldsymbol{\mu}}_m \label{barnu}\\
{\bf R}_{m,d_i}&=&(1+\tilde{\alpha}_i^2 d_i){\bf C}+\tilde{\alpha}_i^2 d_i\sum_{q=1}^M p_{m,q}  (\boldsymbol{\mu}_q-\bar{\boldsymbol{\mu}}_m)(\boldsymbol{\mu}_q-\bar{\boldsymbol{\mu}}_m)^T \label{Rm}
\end{eqnarray}
where $\bar{\boldsymbol{\mu}}_m=\sum_{\ell=1}^Mp_{m,\ell}\boldsymbol{\mu}_\ell$. 
While $\boldsymbol{z}_i$ is non-Gaussian even under the HG, except in the purely homophilic and heterophilic cases, linear processing in the single-layer MLP classifier produces outputs that are nearly Gaussian due to the central limit theorem when $F$ is sufficiently large. Consequently, the first and second-order statistics of $\boldsymbol{z}_i$ are sufficient for approximating the optimum linear classifier and analyzing its performance. Further, since the ${\bf R}_{m,d_i}$'s generally differ across classes, and in order to maintain the classifier's linearity as in single-layer MLP, we use the following pooled covariance matrix: $\Bar{\bf R}_{d_i}=\sum_{m=1}^M \pi_{m,d_i} {\bf R}_{m,d_i}$.   Thus, we consider the following classifier:
\begin{equation}
    \hat{y}_i= \arg\max_m\boldsymbol{w}_{m,d_i}^T\boldsymbol{z}_i+b_{m,d_i}, ~  \boldsymbol{w}_{m,d_i}={\Bar{\bf R}}_{d_i}^{-1}\Bar{\boldsymbol{\nu}}_{m,d_i}, ~ b_{m,d_i}=-\frac{1}{2} \Bar{\boldsymbol{\nu}}_{m,d_i}^T{\bar{\bf R}}_{d_i}^{-1}\Bar{\boldsymbol{\nu}}_{m,d_i}+\log\pi_{m,d_i}, \label{classifierg}
\end{equation}

In practice, estimating the $\pi_{m,d_i}$ values can be challenging due to the limited number of labeled nodes. Therefore, these values may be substituted with their node-degree independent counterparts, the $\pi_{m}$'s, or even with $1/M$. This substitution is unlikely to significantly impact the performance of the resulting classifier unless there is a considerable imbalance in class probabilities.

The deflection coefficients associated with the above classifier can therefore be computed as \begin{equation}
    \gamma_{d_i}(m,\ell)=(\bar{\boldsymbol{\nu}}_{m,d_i}-\bar{\boldsymbol{\nu}}_{\ell,d_i})^T\bar{\bf R}_{d_i}^{-1}(\bar{\boldsymbol{\nu}}_{m,d_i}-\bar{\boldsymbol{\nu}}_{\ell,d_i}). \label{gmk-def}
\end{equation}
The error probability is for nodes having degree  $d_i$ upper bounded by the RHS expression of (\ref{e0m}) after substituting $\gamma_{d_i}(m,\ell)$ for $\gamma_0(m,\ell)$, and $\pi_{m,d_i}$ for $\pi_{m}$.

It is worth pointing out that except in the purely homophilic case, where $\tilde{\alpha}_i=1$ is optimum, there is no normalisation factor to apply to $\boldsymbol{z}_i$ to make the MLP parameters (i.e. the weight vectors and biases) node degree-independent even if the terms related to the a priori probabilities are ignored. Further, the optimised value of $\tilde{\alpha}_i$ is generally node degree dependent for $M>2$, as illustrated in Figure (\ref{perf_M4_gmin4}). Hence, setting $\tilde{\alpha}_i$ and the MLP parameters fixed for all node degrees, as in numerous benchmark GNN models, may undermine performance. 

We consider the following special cases to further illustrate the impact of $\tilde{\alpha}_i$ on performance; simulation results in these special cases will be described in section \ref{simulation}.

\underline{Special case 1}: $M=2$, $p_{1,1}=p_{2,2}=p_h$. \\ In this case, the deflection coefficient and the value of $\Tilde{\alpha}$ that maximises it are respectively found to be 
\begin{equation}
\gamma_{d_i}=\frac{(1+\tilde{\alpha}_i d_i (2p_h-1))^2}{1+d_i  \tilde{\alpha}_i^2+d_i  \tilde{\alpha}_i^2p_h(1-p_h)\gamma_0}\gamma_0, \qquad \tilde{\alpha}_i^*=\frac{2p_h-1}{1+p_{h}(1-p_{h}) \gamma_0}.\label{alpha*}
\end{equation}
The proof is given in Appendix B. 

Unlike the purely homophilic graph case, the graph gain now depends on the graph-agnostic deflection coefficient, and so does the optimal value of $\Tilde{\alpha}_i$. Further, the way  $\Tilde{\alpha}_i^*$ behaves with respect to $p_h$ and $\gamma_0$ provides valuable insight. If $p_h=0.5$, indicating that a neighbor is equally likely to belong to either class, the optimal choice is $\tilde{\alpha}_i^*=0$, suggesting that neighbors should not be utilized for node classification. Conversely, if $p_h=1$ or $p_h=0$, the optimal values are $\tilde{\alpha}_i^*=1$ and $\tilde{\alpha}_i^*=-1$, respectively, as previously determined. Further, it is insightful to analyze the behavior of $\tilde{\alpha}_i^*$ with respect to $\gamma_0$. The value of $\tilde{\alpha}_i^*$ decreases as $\gamma_0$ increases, implying that when the overlap between the distributions of the node feature vectors decreases (i.e., a higher value of $\gamma_0$), there should be less reliance on the neighbors, as involving them introduces uncertainty. Conversely, if $\gamma_0$ is low, the statistical risk of involving neighbors is justifiable. This is illustrated in Figure \ref{perf_M2}.

\underline{Special case 2}: $M=4$, $p_{m,m}=p_h, \forall m$,  $p_{1,2}=p_{2,1}=p_{3,4}=p_{4,3}=1-p_h$.\\
Here, the graph exhibits for all classes either homophily or heterophily, and edges are possible only between classes 1 and 2 and between classes 3 and 4. The deflection coefficients can be computed using  (\ref{gmk-def}). Although the value of $\tilde{\alpha}_i$ that minimises the lower bounds on error probability cannot be obtained in closed-form as in special case 1, it can be easily computed with a one-dimensional grid search. 

\underline{Special case 3}: $M=4$, $p_{1,1}=p_{2,2}=p_{3,4}=p_{4,3}=p_h$, $p_{1,2}=p_{2,1}=p_{3,3}=p_{4,4}=1-p_h$.\\
Here, the graph exhibits a mixture of homophily and heterophily. The same comments made in special case 2 apply.

The aggregation in Eq. (\ref{ave0}), implies that the same processing is applied to the node's feature vector and to those of its neighbors, up to a scalar. More specifically, the same weighting vectors will be applied to the focal node's feature vector and to those of its neighbors. 
Next, we consider another neighborhood aggregation which yields better performance. 

\section{Sum-then-concatenate aggregation}

We consider the aggregation where we first sum the neighbors' feature vectors and then concatenate the result with the focal node's feature vector, i.e. we compute the following ($2F\times 1$) vector:
\begin{equation}
\boldsymbol{z}_i= \left[ \begin{array}{cc}
     \boldsymbol{x}_i  \\
     \boldsymbol{s}_i 
     \end{array}
\right],
 \qquad \boldsymbol{s}_i=\sum_{j\in{\mathcal N}_i} \boldsymbol{x}_j
 \label{AggConc}
\end{equation}
We refer to this as the sum-then-concatenate aggregation (SCA).
This implies that two different sets of weighting vectors may be applied to $\boldsymbol{x}_i$ and $\boldsymbol{s}_i$. This type of GNN is refereed to in \cite{Hamilton-book} as basic GNN message passing. 

Conditioned on $\boldsymbol{n}_i$, the mean and covariance matrix of $\boldsymbol{s}_i$ are respectivelly given by $\boldsymbol{\eta}_m(\boldsymbol{n}_i)=\sum_{q=1}^M n_{i,q} \boldsymbol{\mu}_q
$ and $d_i{\bf C}$.
The unconditional mean and covariance matrix of $\boldsymbol{z}_i$ are:
\begin{eqnarray}
\bar{\boldsymbol{\nu}}_{m,d_i}= \left[ \begin{array}{cc}
  \boldsymbol{\mu}_m  \\
d_i\bar{\boldsymbol{\mu}}_m 
     \end{array}
\right],
 \qquad {\bf R}_{m,d_i}=\left[ \begin{array}{cccc}
  {\bf C} & {\bf 0}  \\
  {\bf 0} & d_i{\bf C}+d_i\sum_{q=1}^M p_{m,q}  (\boldsymbol{\mu}_q-\bar{\boldsymbol{\mu}}_m)(\boldsymbol{\mu}_q-\bar{\boldsymbol{\mu}}_m)^T
     \end{array}
\right],
\end{eqnarray}

We compute the pooled covariance matrix of $\boldsymbol{z}_i$ as in the previous section. Our linear classifier for the SCA approach can thus be expressed as in (\ref{classifierg}) but using the above mean vectors and covariance matrices. It can be rewritten as: 
\begin{equation}
    \hat{y}_i= \arg\max_m\boldsymbol{\mu}_{m}^T{\bf C}^{-1}\boldsymbol{x}_i+\bar{\boldsymbol{\mu}}_{m}^T{\Bar{\bf H}}^{-1}\boldsymbol{s}_i+b_{m,d_i}
\end{equation}   
where
\begin{eqnarray}
    \bar{\bf H}&=&{\bf C}+\sum_{m',q=1}^M\pi_{m',d_i} p_{m',q}  (\boldsymbol{\mu}_q-\bar{\boldsymbol{\mu}}_{m'})(\boldsymbol{\mu}_q-\bar{\boldsymbol{\mu}}_{m'})^T
 \\
     b_{m,d_i}&=&
     -\frac{1}{2} {\boldsymbol{\mu}}_{m}^T{\bf C}^{-1}{\boldsymbol{\mu}}_{m}
      -\frac{1}{2} d_i\Bar{\boldsymbol{\mu}}_{m}^T{\bar{\bf H}}^{-1}\Bar{\boldsymbol{\mu}}_{m}+\log\pi_{m,d_i}
\end{eqnarray}

As for WSA,  even if we ignore the dependency of the a priori probabilities on the $d_i$, the bias term depends on $d_i$; it is an affine function of $d_i$. Normalizing the aggregation of neighbors' feature vectors using $\boldsymbol{s}_i=\frac{1}{d_i}\sum_{j\in{\mathcal N}_i} \boldsymbol{x}_j$, would lead to the following classifier:
\begin{equation}
    \hat{y}_i= \arg\max_m\boldsymbol{\mu}_{m}^T{\bf C}^{-1}\boldsymbol{x}_i+\frac{1}{d_i}\bar{\boldsymbol{\mu}}_{m}^T{\Bar{\bf H}}^{-1}\boldsymbol{s}_i+b_{m,d_i}
\end{equation}   
where
\begin{equation}
     b_{m,d_i}=
     -\frac{1}{2} {\boldsymbol{\mu}}_{m}^T{\bf C}^{-1}{\boldsymbol{\mu}}_{m}
      -\frac{1}{2d_i} \Bar{\boldsymbol{\mu}}_{m}^T{\bar{\bf H}}^{-1}\Bar{\boldsymbol{\mu}}_{m}+\log\pi_{m,d_i}
\end{equation}
Thus, the above-mentioned normalization does not make the optimal linear classifier node degree-dependent. Therefore, as for WSA, using an MLP-based GNN with fixed weighting matrices and bias terms may undermine performance. 

The deflection coefficients defined in (\ref{gmk-def}) can be written as
\begin{equation}
    \gamma_{d_i}(m,\ell)=\gamma_{0}(m,\ell)+d_i(\bar{\boldsymbol{\mu}}_m-\bar{\boldsymbol{\mu}}_\ell)^T\bar{{\bf H}}^{-1}(\bar{\boldsymbol{\mu}}_m-\bar{\boldsymbol{\mu}}_{\ell})
\end{equation}

Consider Special case 3 with $p_h=1$ for which the WSA approach performs poorly for small values of $d_i$; see Figure (\ref{perf_M4_gmin4}). Using the SCA method, the deflection coefficients simplify to 
\begin{eqnarray}
    \gamma_{d_i}(1,2)&=&(1+d_i)\gamma_{0}(1,2), ~  \gamma_{d_i}(3,4)=(1+d_i)\gamma_{0}(3,4), \nonumber\\ 
    \gamma_{d_i}(m,3)&=&\gamma_{0}(m,3)+d_i\gamma_{0}(m,4), \gamma_{d_i}(m,4)=\gamma_{0}(m,4)+d_i\gamma_{0}(m,3), ~ m=1,2
\end{eqnarray}
Hence, with the SCA approach, the deflection coefficients reach their maximum values under Special case 3, unlike with the WSA approach. If the graph agnostic deflection coefficients are all equal to each other, the resulting graph gain  would be equal to $(d_i+1)$.

\section{Simulation results}\label{simulation}
We consider graphs generated using assumptions (A1) and (A2) and the HG scenario, with ${\bf C}=\sigma^2 {\bf I}_F$, $F=20$, and $N=500,000$. We simulate special cases 1, 2, and 3.

For Special case 1, Figure \ref{perf_M2} depicts the minimum error probability of SCA and WSA alongside the corresponding optimal value of $\tilde{\alpha}_i$ for $\gamma_0=1$ and $\gamma_0=10$. The figure also shows the theoretical value of $\tilde{\alpha}_i$, given by Eq. (\ref{alpha*}), which is applicable to Special case 1. The results confirms the finding in (\ref{alpha*}) that the optimum value of $\tilde{\alpha}_i$ is unaffected by the node degree, $d_i$, under assumption (A2).

Figure (\ref{perf_M4_gmin4}) depicts the results for Special cases 2 and 3, and assuming $\gamma_{0}(m,\ell)=4,\forall m\neq \ell$. The WSA-based classifier generally exhibits lower performance compared to the SCA-based classifier, except in scenarios where the graph displays strong homophily, or strong heterophily provided the node degree is not too small. The performance gap is particularly pronounced when the graph exhibits homophily with respect to some classes and heterophily with respect to others, as in Special case 3.

\section{Extensions}\label{extensions}
\subsection{Higher-order neighborhood}\label{subsection:HON}

As mentioned earlier, \cite{Wu2019} has shown that in GNNs, the nonlinearity between layers is not essential. Therefore, we can extend our statistical approach to include higher-order neighbors in a straightforward manner. To illustrate this, we use the SCA approach, where $\boldsymbol{z}_i$ is given by 
\begin{equation}
\boldsymbol{z}_i= \left[ \begin{array}{ccc}
     \boldsymbol{x}_i  \\
     \boldsymbol{s}_i^{(1)} \\
     \vdots \\
     \boldsymbol{s}_i^{(K)}
     \end{array}
\right],
 \qquad \boldsymbol{s}_i^{(k)}=\sum_{j\in{\mathcal N}_i^{(k)}} \boldsymbol{x}_j,\quad k=1,\cdots,K
 \label{AggConcK}
\end{equation}
where $K$ is the number of layers and $\mathcal{N}_i^{(k)}$ is the set of $k$th-order neighbors of node $i$. 

The label-to-label correlations between a node and its $k$th-order neighbors can be approximated using assumption (A2) and by treating equally neighbors of the same order, i.e. the class transition probabilities depend only on the geodesic distances. This results in the following $k$th-order  class transition probability matrix \cite{Hafidi2023}: ${\bf P}^{(k)}={\bf P}^k$ where ${\bf P}$ is the first-order class transition probability matrix whose elements are the $p_{m,\ell}$'s and the elements of ${\bf P}^{(k)}$ are defined as $p_{m,\ell}^{(k)}
\triangleq {\rm Pr}\left( y_j=\ell | y_i=m\right)$ with $j\in \mathcal{N}_i^{(k)}$. 

The classifier in (\ref{classifierg}) and the deflection coefficients can thus be extended to the $K>1$ case in a straightforward manner, as described next.
Our linear  classifier can be expressed as
\begin{equation}
    \hat{y}_i= \arg\max_m\boldsymbol{\mu}_{m}^T{\bf C}^{-1}\boldsymbol{x}_i+\sum_{k=1}^K \bar{\boldsymbol{\mu}}_{m,k}^T{\Bar{\bf H}}_k^{-1}\boldsymbol{s}_i^{(k)}+b_{m,d_i}
\end{equation}   
where
\begin{eqnarray}
    \bar{\bf H}_k&=&{\bf C}+\sum_{k=1}^K\sum_{m',q=1}^M\pi_{m',d_i} p_{m',q}^{(k)}  (\boldsymbol{\mu}_q-\bar{\boldsymbol{\mu}}_{m',k})(\boldsymbol{\mu}_q-\bar{\boldsymbol{\mu}}_{m',k})^T
 \\
     b_{m,d_i}&=&
     -\frac{1}{2} {\boldsymbol{\mu}}_{m}^T{\bf C}^{-1}{\boldsymbol{\mu}}_{m}
      -\sum_{k=1}^K\frac{1}{2} d_{i,k}\Bar{\boldsymbol{\mu}}_{m,k}^T{\bar{\bf H}}_k^{-1}\Bar{\boldsymbol{\mu}}_{m,k}+\log\pi_{m,{\bf d}_i}
\end{eqnarray}
where $\bar{\boldsymbol{\mu}}_{m,k}=\sum_{\ell=1}^Mp_{m,\ell}^{(k)}\boldsymbol{\mu}_\ell$, ${\bf d}_i=\{d_1,d_2,\cdots,d_K\}$ with $d_{i,k}= |\mathcal{N}_i^{(k)}|$ being the number of $k$th-order numbers, and $\pi_{m,{\bf d}_i}$ is the apriori probability that a node belonging to class $m$ has its neighborhood characterised in terms of number of neighbors by ${\bf d}_i$. It is worth noting that in practice,  $K=2$ or $K=3$.    

\subsection{Node-dependent label-to-label correlations}
When the label-to-label correlations signficantly depend on the degrees of the connected nodes, the following broader assumption may be used instead of assumption (A2):

{\bf (A2')}: {\em for any given node, say $i$, of class $m$ and degree $d_i$, the likelihood that a randomly chosen neighbor, say $j$ of degree $d_j$, belongs to class $\ell$ is constant throughout the graph; we denote this probability by $p_{m,\ell}(d_i,d_j)$, i.e.
\begin{equation}
    p_{m,\ell}(d_i,d_j)\triangleq{\rm Pr}\left( y_j=\ell | y_i=m; |\mathcal{N}_i|=d_i;|\mathcal{N}_j|=d_j\right) \nonumber
\end{equation}
}

Graphs under assumption {\bf (A2')} can be generated using a modified version of the degree-corrected version of the stochastic blockmodel \cite{Karrer}, where the likelihood of homophily between two nodes is made to depend on their degrees.

The derivations obtained under assumption (A2) should be reexamined in light of assumption (A2'). For instance, while assigning equal weights to neighbors is logical under assumption (A2), assumption (A2') suggests that allowing these weights to vary based on the degrees of neighboring nodes could potentially enhance performance. This however introduces complexities in parameter optimization, which could be alleviated if the $p_{m,\ell}(d_i,d_j)$'s are smooth and monotonic functions of $d_i$ and $d_j$.    

\section{Discussion and Conclusions}

In this paper, we have critically examined the issue of GNN-based node classification under the assumption of edge-independent node labels, with a particular emphasis on reevaluating neighborhood aggregation. By adopting a statistical signal processing perspective, we have demonstrated that GNN models, such as those based on graph spectral convolution and graph isomorphism tests, have limitations under this assumption. We constructed linear classifiers and investigated two types of linear aggregations: weighted sum aggregation (WSA) and sum-and-concatenate aggregation (SCA). Our results indicate that the latter may significantly outperform the former, suggesting that the focal node's features and the aggregated neighbors' feature vector should undergo different processing.

The superiority of SCA over WSA was demonstrated assuming complete knowledge of the graph's statistical properties, particularly the feature vectors' distributions and the label-to-label correlation matrix. This assumption allowed us to analytically derive the weight vectors and biases of the optimal linear classifiers. However, in practical scenarios, these statistical parameters must be estimated using supervised or semi-supervised methods. When the dimension of the feature vector, $F$, is large, reducing the number of parameters through simplifying assumptions may be beneficial. For example, the correlations between the features may be ignored, i.e. the feature covariance matrix ${\bf C}$ can be assumed to be diagonal, as in the Naive Bayes classifier. Therefore, it is crucial to investigate the impact of estimation inaccuracies and any modeling simplifications on the performance of WSA and SCA before drawing definitive conclusions in any specific context. 

Furthermore, from a practical standpoint, the derived optimal linear classifiers should be compared to the theoretically suboptimal, unstructured single-layer MLP classifiers used in conventional GNN, i.e. $\hat{y}_i= \arg\max_m\boldsymbol{w}_{m}^T\boldsymbol{z}_i+b_{m}$ for WSA, and $\hat{y}_i= \arg\max_m\boldsymbol{w}_{m,{\rm self}}^T\boldsymbol{x}_i+\boldsymbol{w}_{m,{\rm neigh}}^T\boldsymbol{s}_i+b_{m}$ for SCA, where $\boldsymbol{w}_m$, $\boldsymbol{w}_{m,{\rm self}}$ and $\boldsymbol{w}_{m,{\rm neigh}}$, $m=1\cdots,M$, are fixed unstructured vectors to optimize by minimizing an empirical risk (e.g. cross entropy). 
The latter classifiers have fewer parameters to estimate, potentially offering greater robustness when the estimation of graph parameters is not sufficiently accurate.

Overall, this study underscores the importance of a thorough conceptual understanding of neighborhood aggregation in GNNs and opens avenues for the design of GNN models that achieve superior performance. We hope that our work stimulates further research in this direction, ultimately contributing to the advancement of graph-based learning and its applications.

\bibliography{main}


\vspace{1cm}
\section*{Appendices}
{\bf Appendix A}:
Classification error probability of the graph-agnostic classifier.

In general, the classification error probability is defined and expressed as: 
\begin{equation}
    e\triangleq{\rm Pr}(\hat{y}_i\ne y_i) = \sum_{m=1}^M \pi_m {\rm Pr}(\hat{y}_i\ne m | y_i=m)
\end{equation}

For the graph-agnostic classifier in Subsection \ref{GACsubsection}, and under the HG scenario, ${y}_i\ne m$ when $y_i=m$ if there exists an $\ell$ such that $T_{i,\ell} > T_{i,m}$, where $T_{i,\ell}= \boldsymbol{\mu}_\ell^T{\bf C}^{-1}\boldsymbol{x}_i-(1/2)\boldsymbol{\mu}_\ell^T{\bf C}^{-1}\boldsymbol{\mu}_\ell+\log(\pi_\ell)$, with $\boldsymbol{x}_i$ being a random vector generated using $G(\boldsymbol{x}_i | \boldsymbol{\mu}_m,{\bf C})$. That is, for $m=1$, we have that
\begin{equation}
{\rm Pr}(\hat{y}_i\ne 1 | y_i=1)={\rm Pr}(T_{i,2}> T_{i,1}~{\rm or}~  \cdots ~{\rm or} ~T_{i,M}>T_{i,1} | y_i=1) 
\end{equation}
The same applies for other values of $m$.
Using the probability rule ${\rm Pr}(A ~{\rm or}~ B)\le {\rm Pr}(A) + {\rm Pr}(B)$, we obtain the following inequalities:
\begin{equation}
{\rm Pr}(\hat{y}_i\ne 1 | y_i=m) \le \sum_{\ell\ne m} {\rm Pr}(T_{i,\ell}>T_{i,m} | y_i=m), \qquad m=1,\cdots,M
\end{equation}
We next derive ${\rm Pr}(T_{i,\ell}>T_{i,m} | y_i=m)$. 

When $y_i=m$, $\boldsymbol{x}_i$ can be modeled as $\boldsymbol{\mu}_m+{\bf C}^{1/2} \boldsymbol{\epsilon}_i$ with $\boldsymbol{\epsilon}_i$ being a standard Gaussian $(F\times 1)$ vector. Thus, $T_{i,\ell}>T_{i,m}$ if 
\begin{equation}
    (\boldsymbol{\mu}_\ell-\boldsymbol{\mu}_m)^T{\bf C}^{-1/2}{\bf \epsilon} >\frac{1}{2}(\boldsymbol{\mu}_\ell-\boldsymbol{\mu}_m)^T{\bf C}^{-1}(\boldsymbol{\mu}_\ell-\boldsymbol{\mu}_m)+\log\frac{\pi_m}{\pi_\ell}
\end{equation}
The RHS of the above inequality is a zero-mean Gaussian variable with variance equal to $(\boldsymbol{\mu}_\ell-\boldsymbol{\mu}_m)^T{\bf C}^{-1}(\boldsymbol{\mu}_\ell-\boldsymbol{\mu}_m)$.
It follows that 
\begin{equation}
     {\rm Pr}(T_{i,\ell}>T_{i,m} | y_i=m)={\rm Pr}\left(\epsilon > \frac{\sqrt{\gamma_{m,\ell}}}{2}+ \frac{1}{\sqrt{\gamma_{m,\ell}}}\log\frac{\pi_m}{\pi_\ell}\right)
\end{equation}
where $\epsilon$ is a standard Gaussian random variable and $\gamma_0(m,k)=(\boldsymbol{\mu}_m-\boldsymbol{\mu}_\ell)^T{\bf C}^{-1}(\boldsymbol{\mu}_m-\boldsymbol{\mu}_\ell)$, with $m\neq \ell$. This probability can be evaluated using the $Q$-function as shown in Equation (\ref{e0m}). 

\vspace{0.5cm}

{\bf Appendix B}: Deflection coefficient in Special Case 1.

If $M=2$, the covariance matrix of $\boldsymbol{z}_i$ in Eq. (\ref{ave1}) is
\begin{equation}    
\bar{\bf R}_{d_i}=(1+ \tilde{\alpha}_i^2d_i){\bf C}+ 
 \tilde{\alpha}_i^2 d_i\sum_{\ell=1}^2  p_{m,\ell}(\boldsymbol{\mu}_\ell-\bar{\boldsymbol{\mu}}_m)(\boldsymbol{\mu}_\ell-\bar{\boldsymbol{\mu}}_m)^T 
\end{equation}
where we recall that $\bar{\boldsymbol{\mu}}_m=\sum_{\ell=1}^Mp_{m,\ell}\boldsymbol{\mu}_\ell$. If $p_{1,1}=p_{2,2}=p_h$, this  covariance matrix can be rewritten as:
\begin{equation}    
\bar{\bf R}_{d_i}=(1+d_i  \tilde{\alpha}_i^2){\bf C}+ 
d_i  \tilde{\alpha}_i^2 \sum_{\ell=1}^2  p_{1,\ell}(\boldsymbol{\mu}_\ell-p_{h}\boldsymbol{\mu}_1-(1-p_{h})\boldsymbol{\mu}_2)
(\boldsymbol{\mu}_\ell-(1-p_{h})\boldsymbol{\mu}_1-p_{h}\boldsymbol{\mu}_2)^T 
\end{equation}
After straightforward derivations, we obtain the following expression:
\begin{equation}    
\bar{\bf R}_{d_i}(\tilde{\alpha}_i)=(1+d_i  \tilde{\alpha}_i^2){\bf C}+ 
d_i  \tilde{\alpha}_i^2   p_h(1-p_{h})(\boldsymbol{\mu}_1-\boldsymbol{\mu}_2)
(\boldsymbol{\mu}_1-\boldsymbol{\mu}_2)^T 
\end{equation}

Using the Sherman–Morrison formula, we obtain the following closed-form expression for the inverse of $\bar{{\bf R}}_{d_i}$:
\begin{equation}
    \bar{\bf R}_{d_i}^{-1}=\frac{1}{1+  \tilde{\alpha}_i^2d_i} \left({\bf C}^{-1}-\xi {\bf C}^{-1} (\boldsymbol{\mu}_1-\boldsymbol{\mu}_2)(\boldsymbol{\mu}_1-\boldsymbol{\mu}_2)^T
    {\bf C}^{-1} \right )
\end{equation}
where
\begin{equation}
    \xi=\frac{  \tilde{\alpha}_i^2d_ip_h(1-p_h)}{1+  \tilde{\alpha}_i^2d_i+\tilde{\alpha}_i^2d_ip_h(1-p_h)\gamma_0}
\end{equation}

Further, we have that 
$(\bar{\boldsymbol{\nu}}_{1,d_i}-\bar{\boldsymbol{\nu}}_{2,d_i})=(1+\tilde{\alpha}_i d_i (2p_h-1)) (\boldsymbol{\mu}_1-\boldsymbol{\mu}_2)$

So, the deflection coefficient can thus be expressed as

\begin{equation}
\gamma_{d_i}=\frac{(1+\tilde{\alpha}_i d_i (2p_h-1))^2(1-\xi \gamma_0)}{1+  \tilde{\alpha}_i^2d_i} \gamma_0
\end{equation}

After some straightforward derivations, we obtain the expressions for the deflection coefficient and the optimal value of $\Tilde{\alpha}$ in Eq. (\ref{alpha*}). 

\newpage

 \end{document}